\documentclass[lettersize,journal]{IEEEtran}
\usepackage{amsmath,amsfonts}
\usepackage{mathabx}
\usepackage[colorlinks,linkcolor=blue,anchorcolor=blue,citecolor=blue,urlcolor=blue]{hyperref}
\usepackage[linesnumbered, ruled]{algorithm2e}

\providecommand{\DontPrintSemicolon}{\dontprintsemicolon}
\usepackage{array}
\usepackage{textcomp}
\usepackage{stfloats}
\usepackage{url}
\usepackage{verbatim}
\usepackage{graphicx}
\usepackage{cite}
\usepackage{lipsum}
\usepackage{multirow}
\usepackage[caption=false,font=footnotesize]{subfig}
\usepackage{threeparttable}

\usepackage{tikz}
\usetikzlibrary {arrows.meta,automata,positioning,shadows}
\usepackage{pgfplots}
\pgfplotsset{compat=1.18}

\newcommand{\mc}[1]{\mathcal{#1}}

\begin{document}

\title{GPLight+: A Genetic Programming Method for Learning Symmetric Traffic Signal Control Policy}

\definecolor{lime}{HTML}{A6CE39}
\DeclareRobustCommand{\orcidicon}{
    \hspace{-2.5mm}
    \begin{tikzpicture}
    \draw[lime, fill=lime] (0,0)
    circle [radius=0.16]
    node[white] {{\fontfamily{qag}\selectfont \tiny ID}};
    \draw[white, fill=white] (-0.0625,0.095)
    circle [radius=0.007];
    \end{tikzpicture}
    \hspace{-2mm}
}

\foreach \x in {A, ..., Z}{%
    \expandafter\xdef\csname orcid\x\endcsname{\noexpand\href{https://orcid.org/\csname orcidauthor\x\endcsname}{\noexpand\orcidicon}}
}

\newcommand{\orcidauthorA}{0009-0001-6992-8339}
\newcommand{\orcidauthorB}{0000-0003-0682-1363}
\newcommand{\orcidauthorC}{0000-0003-4463-9538}

\author{Xiao-Cheng Liao\orcidA{}, \and
Yi Mei\orcidB{}, \textit{Senior Member, IEEE,} \and
Mengjie Zhang\orcidC{}, \textit{Fellow, IEEE}

\IEEEcompsocitemizethanks{\IEEEcompsocthanksitem \hyphenpenalty=8000 The authors are with the Centre for Data Science and Artificial Intelligence \& School of Engineering and Computer Science, Victoria University of Wellington, Wellington 6140, New Zealand (E-mail: xiaocheng@ecs.vuw.ac.nz; yi.mei@ecs.vuw.ac.nz; mengjie.zhang@ecs.vuw.ac.nz).

This article has supplementary downloadable material available at https://doi.org/10.1109/TEVC.2025.3578575, provided by the authors.
}
}

\markboth{Journal of \LaTeX\ Class Files,~Vol.~14, No.~8, August~2024}%
{XXX \MakeLowercase{\textit{et al.}}: A Sample Article Using IEEEtran.cls for IEEE Journals}


\maketitle
 
\begin{abstract}
Recently, learning-based approaches, have achieved significant success in automatically devising effective traffic signal control strategies.
In particular, as a powerful evolutionary machine learning approach,
Genetic Programming (GP) is utilized to evolve human-understandable phase urgency functions to measure the urgency of activating a green light for a specific phase.
However, current GP-based methods are unable to treat the common traffic features of different traffic signal phases consistently.
To address this issue, we propose to use a symmetric phase urgency function to calculate the phase urgency for a specific phase based on the current road conditions.
This is represented as an aggregation of two shared subtrees, each representing the urgency of a turn movement in the phase. 
We then propose a GP method to evolve the symmetric phase urgency function.
We evaluate our proposed method on the well-known cityflow traffic simulator, based on multiple public real-world datasets.
The experimental results show that the proposed symmetric urgency function representation can significantly improve the performance of the learned traffic signal control policies over the traditional GP representation on a wide range of scenarios.
Further analysis shows that the proposed method can evolve effective, human-understandable and easily deployable traffic signal control policies.

\end{abstract}


\section{Introduction}
\IEEEPARstart{T}{raffic} signals, located at signalized intersections, manage traffic flow in various directions, thereby significantly contributing to the improvement of both transportation efficiency and road safety \cite{liao2022combining}.
Poorly designed traffic signal plans result in commuters wasting valuable time on the roads.
The majority of existing traffic signal control systems do not operate based on decisions tailored to the dynamic traffic conditions.
For instance, the Sydney Coordinated Adaptive Traffic System \cite{lowrie1990scats}, which relies on a predetermined cycle time plan, remains extensively utilized in real signalized intersections worldwide.

The emergence of Deep Reinforcement Learning (DRL) as a solution to the Traffic Signal Control (TSC) problem is driven by advancements in deep learning \cite{lecun2015deep} and the increasing accessibility of transportation infrastructure components such as surveillance cameras, road sensors, and the internet of vehicles \cite{ji2020survey}. This trend is exemplified by recent research efforts \cite{chen2020toward,wei2019colight, wei2019presslight}.
DRL techniques enable the exploration of effective traffic signal control strategies by leveraging feedback from the surrounding environment.
Such strategies facilitate the dynamic adjustment of traffic signals in response to real-time conditions at intersections, showcasing superior performance compared to traditional transportation methods \cite{haydari2020deep}.

Despite the promising results achieved in TSC through DRL methods, two major problems persist within the existing frameworks.
One issue is the intricate design required for essential components, with particular emphasis on the demanding nature of the reward system \cite{kwon2023reward, hadfield2017inverse}.
Crafting a meaningful reward often necessitates a substantial amount of domain knowledge and expertise \cite{wei2019presslight}.
Without essential domain knowledge, it is easy to encounter challenges like excessively delayed rewards and difficulties in appropriately assigning credit over extended time scales \cite{salimans2017evolution}.
For example, within a single time step, accurately assessing the effectiveness of an action, such as transitioning from one traffic signal phase 
to another, using an immediate reward is challenging.

The other issue arises from the fact that the signal control policy learned by DRL typically relies on a complex neural network.
This complexity hampers human or expert comprehension, rendering it challenging to interpret and explain the learned policy \cite{ali2023explainable, ghorbani2019interpretation,zhu2022extracting}.
However, the importance of explainable traffic signal control strategies cannot be overstated \cite{zhu2022extracting}.
The lack of transparency poses significant barriers to establishing trust among users and hinders dispatchers from thoroughly examining potential weaknesses in the policies \cite{sheh2017did,mei2022explainable}.
Moreover, drivers should be able to anticipate the next traffic light change while awaiting a green signal; failure to do so may result in traffic confusion.
For instance, drivers in the through lane might abruptly enter the right-turn lane because they are unable to anticipate the arrival of a green signal.

To tackle the aforementioned shortcomings, this paper introduces a traffic signal control optimization algorithm based on genetic programming (GP) \cite{koza1994genetic}.
GP 
has been recently demonstrated to automatically evolve effective traffic signal control policies.
For example, in GPLight \cite{liao2024learning}, a concept of urgency is assigned to each traffic signal phase, and the tree-based urgency function is evolved through GP.
Whenever a transition in traffic signal phases is required at an intersection, the urgency function takes into account the traffic movement characteristics on the lanes that could be impacted by the phase change, thereby generating an urgency value for that particular phase.
The phase urgency generated can be regarded as the priority for a green light demand given the current intersection state. 
As a result, the phase with the highest phase urgency is actuated next.

\begin{figure}[!t]
\centering
\includegraphics[width=\columnwidth]{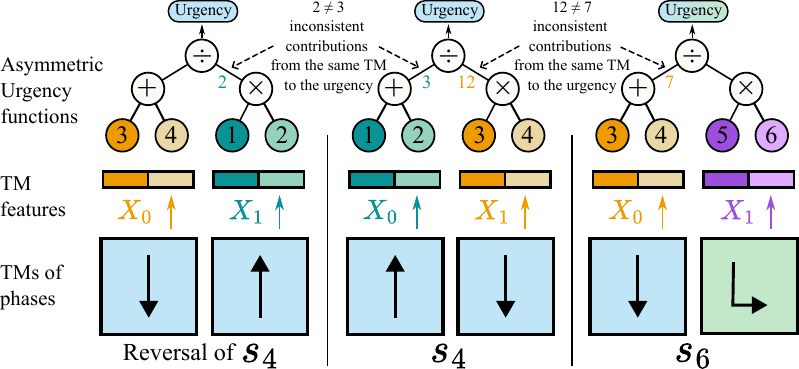}
\caption{An example of calculating the urgency of three different phases (i.e., $s_6$ and $s_4$ with its reversal) using an asymmetric phase urgency function, where white circular nodes represent operators and the circular nodes with different color represent the features values of TMs.
The features of the two TMs involved in a phase serve as inputs of the urgency function, which will then output the phase urgency value.
Different phases share the same TM.
The same TM contributes inconsistently to the phase urgency in different phases.}
\label{fig:inconsistency}
\end{figure}

Although the two turn movements (TMs) involved in a phase are unordered, the existing studies \cite{liao2024learning,zheng2019diagnosing,wei2019presslight,wei2018intellilight} treated them as ordered, and learn policies based on a predefined order. 
In other words, the learned policy will provide a different urgency value for the same phase after simply swapping the order of the two involved TMs. 
Such order dependence not only introduces complexity into the learning process by expanding the input space of urgency function but also reduces decision stability and algorithm performance due to inconsistent urgency evaluations.
An example of calculating the urgency of three different phases using an asymmetric phase urgency function is presented in Fig. \ref{fig:inconsistency}.
By observing the urgency calculations of phases $s_4$ and $s_6$, it can be seen that if the phase urgency function is not symmetric, the same TM might contribute differently to the final urgency across various phases.
Besides, by observing the urgency calculations of $s_4$ and its reversal, we can see that the contributions of these two TMs to the final phase urgency are completely reversed.
This phenomenon is counterintuitive.
In real-world scenarios, if traffic flow undergoes a symmetric change (e.g., during the morning and evening peaks), where the features of $X_0$ and $X_1$ are exchanged, it will be reasonable to expect that the phase urgency would remain unchanged.

To address this issue, we propose to introduce symmetry constraints in the phase urgency function. 
Specifically, we enforce symmetry by restricting subtrees that can be regarded as the urgency function for individual TM, to share the same structure, ensuring that the final phase urgency value remains invariant regardless of the order of the involved TMs.
The main contributions of this paper are as follows:

\begin{enumerate}
    \item We present a unified framework for complete full-phase multi-intersection traffic signal control using GP, which defines a new urgency function representation, a lane feature permutation template and a methodology for determining the next phase based on the urgency function.

    \item We introduce a symmetry constraint to the phase urgency function to ensure that the urgency function will always output the same urgency value for the same phase, regardless of the order of the TMs in it, and the common features are treated consistently across different phases.
    Instead of using GP to optimize the entire phase urgency function, we choose to optimize its subtrees and then assemble them to construct the complete urgency function, ensuring that the tree-based functions evolved by GP always meet the above constraint.
    We demonstrate a significant performance improvement over GPLight on benchmark real-world datasets after imposing the symmetric constraints on the phase urgency function.

    \item 
    We verify the effectiveness of proposed method and analyze the evolved traffic signal control policies, covering aspects such as lane feature importance, explainability, generalization to unseen scenarios, computation time, and memory usage.

\end{enumerate}

The rest of this paper is organized as follows.
In Section \ref{sec:background}, we review the prior studies concerning traffic signal control and introduce the traffic signal control problem.
We present the proposed GPLight+\footnote{Code available at https://github.com/Rabbytr/gplight} in Section \ref{sec:method} and validate its effectiveness and advantages in Section \ref{sec:experiments}.
Finally, we conclude the work in Section \ref{sec:conclusion}.

\section{Background}
\label{sec:background}

\subsection{Problem Definition} 

In this subsection, some important definitions related to the traffic signal control to be solved in this work is introduced.
Based on these definitions, we define the optimization objectives for multi-intersection traffic signal control.

\subsubsection{Road}

A road $\mc{R}_{r}$ typically consists of multiple lanes, and vehicles on these lanes can only travel in the same direction, meaning that the roads are directed.

\subsubsection{Road network}

A road network consists of a number of roads $\{\mc{R}_0, \mc{R}_1, \mc{R}_2, \cdots \}$ and intersections $\{\mc{I}_0, \mc{I}_1, \mc{I}_2, \cdots \}$, interconnected to form a complete transportation system that facilitate vehicular and pedestrian movement within a geographical area.
It is a network with infrastructures designed to efficiently manage and accommodate different traffic flows.

\subsubsection{Intersection}

An intersection is a critical point within the road network where two or more roads meet or cross each other. 
It serves as a focal point for coordinating different traffic flows with different directions.

\begin{figure}[!t]
\centering
\includegraphics[width=0.6\columnwidth]{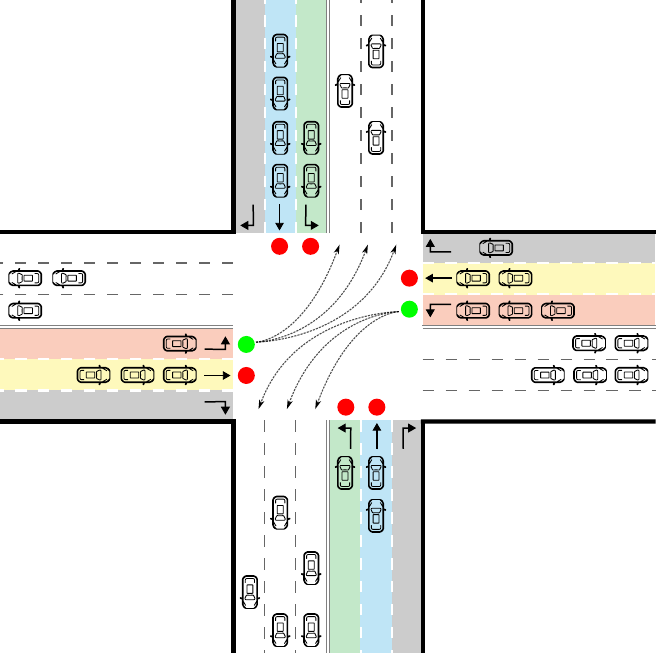}
\caption{An Example of an intersection. The intersection has four incoming roads and four outgoing roads, with the incoming roads colored for clarity.}
\label{fig:intersection}
\end{figure}

An example of an intersection structure is presented in Fig. \ref{fig:intersection}.
It consists of eight roads that can be categorized into two types: incoming roads and outgoing roads.
Each road is composed of three lanes, one turning left, one going straight, and one turning right. 
In total, there are twelve incoming lanes and twelve outgoing lanes in an intersection. 
A vehicle arriving on an incoming lane $l$ can cross the intersection and move to one of the corresponding outgoing lanes $m$.

Intersections are dynamic environments where traffic movements must be carefully managed to ensure safety, efficiency, and compliance with traffic regulations.

\subsubsection{Turn movement}

\begin{figure}[!t]
\centering
\includegraphics[width=0.55\columnwidth]{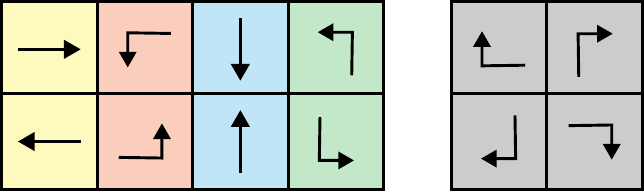}
\caption{Eight different traffic turn movements (right turn is always allowed, thus not considered in the work).}
\label{fig:tms}
\end{figure}

A turn movement (TM) at an intersection refers to the traffic traveling across the intersection from one incoming road to an outgoing road. 
At a single intersection, vehicles from each of the four incoming roads can turn left, go straight, or turn right.
As presented in Fig. \ref{fig:tms}, there are $3 \times 4 = 12$ different types of turn movements, including four left-turn TMs, four go-straight TMs and four right-turn TMs.
Among them, the right turn TMs are assumed to be permitted at all times\footnote{``Right Turn on Red'' a principle of law permitting at a red traffic light to make a turn into the nearest traffic lane without waiting for a green signal. This practice is permitted in many countries, such as China and the United States.}. 
Therefore, there are eight TMs that need to be controlled by traffic signals.

Notably, we assume that traffic must travel on the right side.
However, in some countries, such as UK, Australia and New Zealand, vehicles drive on the left side. 
This situation represents a mirrored setting and does not affect the adaptability of the methods presented in this paper.

\subsubsection{Traffic phase}
\begin{figure}[!t]
\centering
\includegraphics[width=0.65\columnwidth]{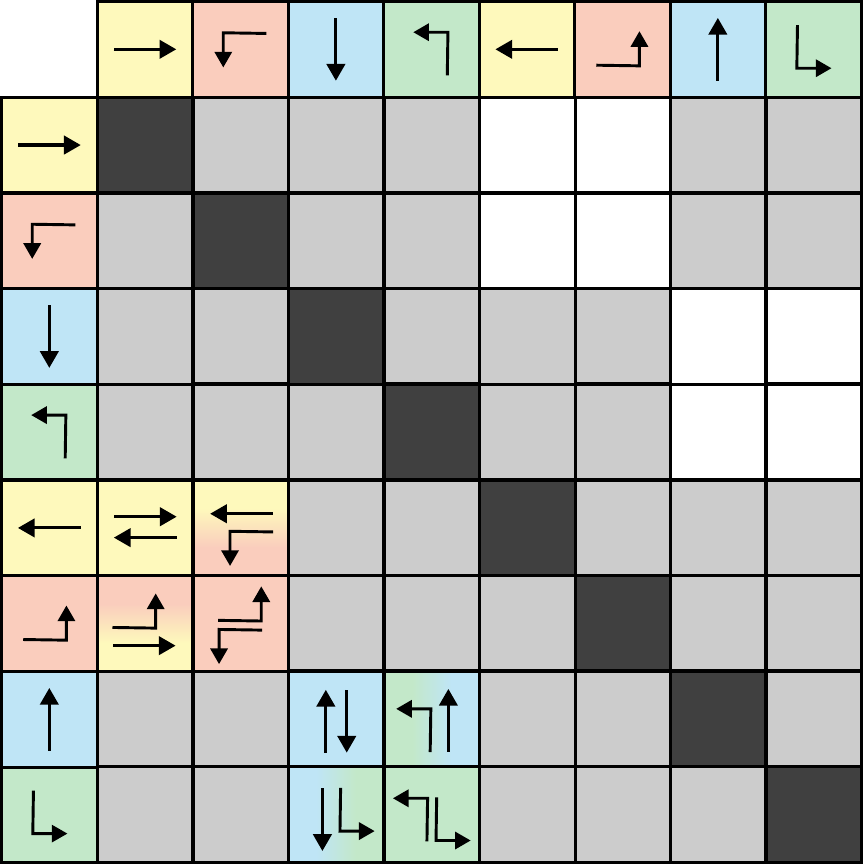}
\caption{Eight traffic phases composed of 8 non-conflicting pairs of turn movements. The eight black cells along the diagonal represent phases that consist of the same TM and are not considered in a standard 8-phase setting.
The eight white cells, symmetrically positioned along the diagonal, represent the symmetric phases among the eight phases, indicating that the order of their TMs is swapped.}
\label{fig:phase_matrix}
\end{figure}

A signal traffic phase at an intersection refers to a pair of permitted TMs.
In a traffic phase, its corresponding permitted TMs are given green light, while all the other TMs are given red light.
Fig. \ref{fig:phase_matrix} illustrates the mutual relationships among all TMs.
In this figure, the gray cell indicates that the traffic flow directions for the corresponding pair of TMs are incompatible. 
Excluding these conflicting pairs, the remaining 8 pairs of TMs that are mutually compatible form 8 different traffic phases.

\subsubsection{Objective}
A traffic network consists of intersections and roads. 
Given the time period of analysis, each intersection is controlled at each step $t$, making an optimal decision to choose appropriate signal phase by observing the current traffic conditions at the intersection.
If the next phase is the same as the current phase, the duration of the current phase will be extended. 
Otherwise, each phase transition is accompanied by a three-second yellow light and a two-second red light to clear the vehicles from the intersection.
The objective is to minimize the average travel time for all vehicles spent between their entering and leaving areas in the network.
Notably, the TSC problem is typically defined with the average travel time of vehicles as the sole objective in many previous works \cite{chen2020toward,liao2024learning,zheng2019learning}.
In fact, some studies have demonstrated that the average travel time has certain consistency with other potential traffic metrics, such as the vehicle queue length \cite{zheng2019diagnosing} and network throughput \cite{wei2019presslight}.
Accordingly, we use the average travel time as the metric.

\subsection{Related Work}

A considerable number of traffic signal control methods have been proposed, categorizable into three typical groups \cite{zheng2019learning}: traditional methods, optimization-based methods and learning-based methods.

\subsubsection{Traditional Traffic Signal Control Methods}
Traffic signal control methods traditionally fall into three distinct subgroups.

\textit{Fixed time control.}
In 1958, Webster \textit{et al.} \cite{webster1958traffic} pioneered the fixed-time control approach, which involves adjusting signal phases based on predetermined rules outlined in signal plans.
The fixed-time method governs the switching of traffic signal phases in a mechanical sequence, disregarding the vehicle states across different lanes.
Such an approach could result in significant traffic flow imbalances. Nonetheless, due to its simplicity, it finds widespread adoption \cite{wu2021efficient} in real-world applications.

\textit{Actuated methods.}
Several studies \cite{fellendorf1994vissim, mirchandani2001real} have established a set of rules, where the activation of traffic signals depends on adherence to these predefined rules along with real-time data.
As an example, a rule might specify that the green signal is allocated to a particular turn movement only when the queue length surpasses a predetermined threshold.

\textit{Pre-defined plan-based methods.}
In subsequent advancements, certain methods and systems \cite{koonce2008traffic}, such as SCATS \cite{lowrie1990scats} and SCOOT \cite{hunt1982scoot}, coordinate traffic signals according to a series of signal plans that are manually designed or automatically designed \cite{keblawi2023automatic}.
In these approaches, a series of signal plans are pre-established, and then, depending on the real-time traffic conditions of the roads, the system determines which signal plan to implement.

\subsubsection{Optimization-based methods}

Conventional methods for traffic signal control rely heavily on human expertise, as they require the manual creation of traffic signal plans or rules.
Also, there is a lack of integration with an optimization process, which can result in potential performance shortcomings.
Classical optimization-based methods usually aim to optimize travel time by assuming a uniform arrival rate \cite{webster1966traffic,roess2004traffic}.
Following this, a traffic signal plan, which includes cycle length and phase ratios, can be computed using formulas derived from traffic data.
Certain researchers employ meta-heuristic algorithms, including genetic algorithms \cite{li2018signal,tung2014novel}, particle swarm optimization \cite{jia2019multi, chuo2017evolvable}, differential evolution \cite{cakici2019differential, baskan2019multiobjective}, and ant colony optimization \cite{baskan2011ant, renfrew2012traffic}, for the optimization of signal plans.
Typically, decision variables such as signal type, cycle time, signal offset, and green time are utilized.
Optimization-based methods rely less on human knowledge, as they determine traffic signal plans based on observed traffic data, and have shown promising results.
Nonetheless, these methods still depend on pre-optimized signal plans, posing challenges in addressing dynamic and uncertain traffic conditions.

Varaiya introduced the Max-pressure (MP) method \cite{varaiya2013max}, which eliminates the need for predefined signal plans and enables the adjustment of traffic signals according to real-time traffic conditions at intersections.
MP stands as a state-of-the-art method in the transportation field \cite{wei2019presslight}, with its primary aim being to maximize network throughput, consequently minimizing travel time.
However, it still operates on strong assumptions, such as assuming unlimited capacity of downstream lanes, which is used to simplify traffic conditions \cite{wei2019presslight}.
Such assumptions may restrict the effectiveness of the MP method in real-world scenarios, as they may not consistently align with actual conditions.

\begin{figure*}[!t]
\centering
\includegraphics[width=0.86\textwidth]{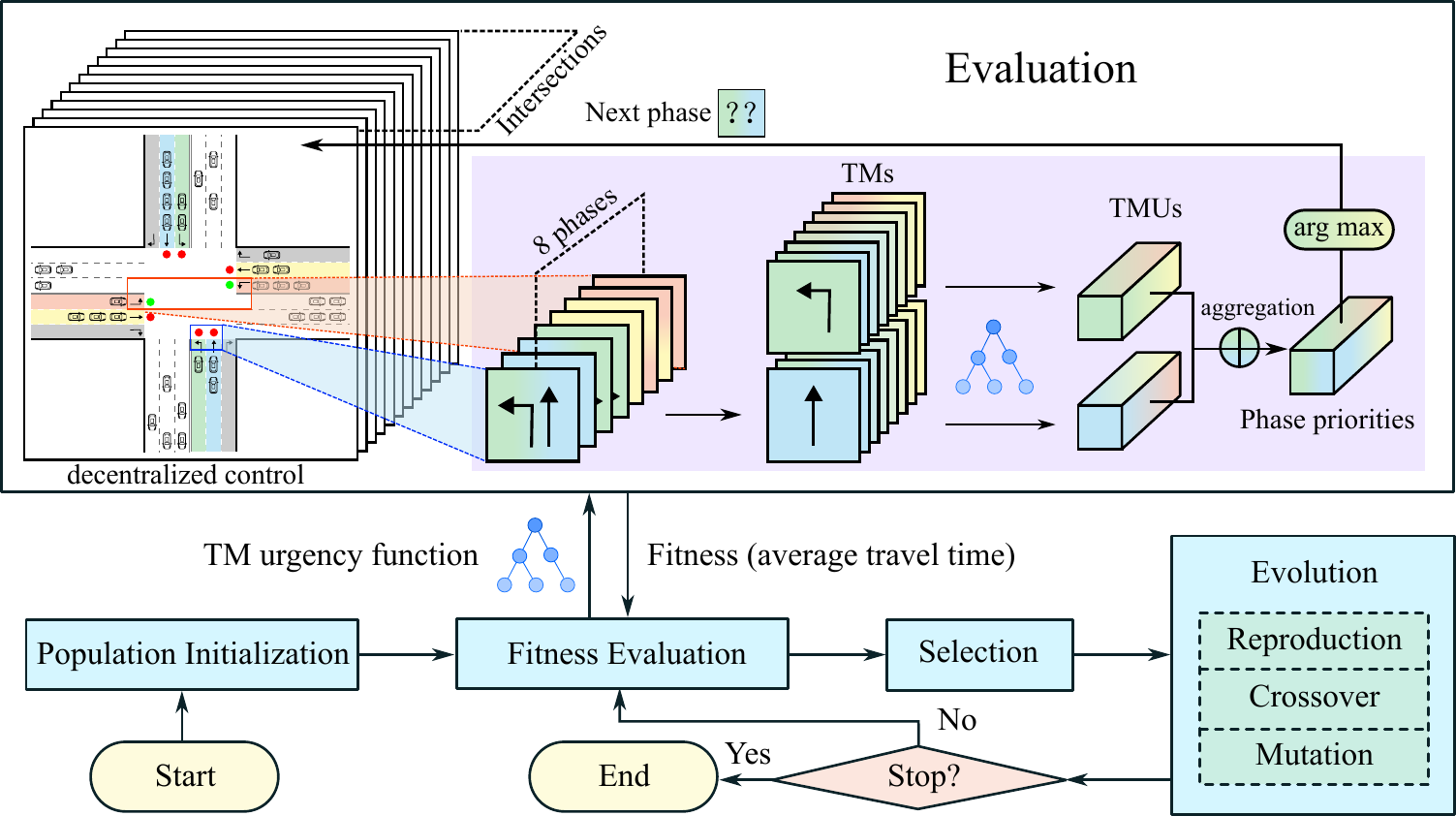}
\caption{The overall framework of the proposed GPLight+ method, with the purple shaded area highlights the primary distinctions between GPLight+ and GPLight.}
\label{fig:framework}
\end{figure*}

\subsubsection{Learning-based methods}

Unlike traditional methods, learning-based approaches do not necessitate predefined or pre-optimized static signal plans \cite{keblawi2023automatic} and refrain from making strong assumptions about traffic data.
Learning-based approaches can directly acquire knowledge from intersections \cite{du2023safelight}, utilizing feedback from the transportation system without prior familiarity with a given environment.

Currently, the most dominant methods for traffic signal control are based on DRL.
Within DRL, a strategy involves centrally managing signals across all intersections within the road network \cite{prashanth2010reinforcement}.
This entails the agent directly making decisions for all intersections, yet mastering this task poses challenges due to the curse of dimensionality in the action space.
Some studies, like \cite{kuyer2008multiagent} and \cite{van2016coordinated}, have delved into employing a multi-agent DRL approach to jointly model two adjacent intersections, utilizing centralized global joint actions.
However, the deployment of these approaches encounters scalability issues.
As the network scale expands, centralized optimization becomes infeasible due to the combinatorially large joint action space. This impedes the widespread adoption of this method for city-level control.

To address this challenge, methods were proposed to model each intersection as an individual agent \cite{wiering2000multi}.
For example, Zheng \textit{et al.} \cite{zheng2019learning} proposed a well-designed neural network model called FRAP based on phase relations and competition.
This model can effectively handle unbalanced traffic flow while maintaining the rotational invariance of traffic flow directions at an intersection.
In their work, a decentralized framework was employed, wherein a set of FRAP-based agents was trained, with each agent responsible for managing a specific intersection.
Wei \textit{et al.} \cite{wei2019presslight} also adopted a decentralized paradigm and designed an RL reward based on the concept of pressure, demonstrating a better-designed reward can lead to significant performance improvements.
Furthermore, MPLight \cite{chen2020toward} uses the difference between upstream and downstream queue lengths, which reflects the imbalance in vehicle distribution, as its reward.
By minimizing this imbalance, the RL agent can balance the vehicle distribution within the system, maximize system throughput, and, in turn, reduce the average travel time of vehicles.
Additionally, MPLight utilizes the network structure of FRAP as the base model for its superior performance and generalizability to enable parameter sharing among different intersections.
This approach demonstrates that having all intersections follow the same traffic signal control policy can significantly enhance both the convergence speed and scalability, and that a well-designed model-agnostic reward can greatly improve the effectiveness of the policy.
However, these methods usually necessitate crafting a reward function that might not directly align with the problem objective, typically the average travel time of vehicles, and they frequently require a certain degree of domain expertise.

Besides RL-based approaches, Genetic Programming (GP) recently showed certain advantages for rules or policies learning \cite{jia2022learning,zhang2023survey,bookzhang2021genetic, nguyen2017genetic}, such as effectiveness and interpretability of the final rules \cite{xu2024niching,xu2024genetic}.
For the TSC problem,
Ricalde and Banzhaf \cite{ricalde2016genetic, ricalde2017evolving} conducted initial investigations utilizing GP with an epigenetics design within the TSC domain.
However, their approach operates in scenarios featuring solely vertical and horizontal traffic flows, akin to a 2-phase setup, which is less common in real-world situations.
Liao \textit{et al.} \cite{liao2024learning} use GPLight to evolve an urgency function to compute the priority of each traffic phase.
Their method can be used in the complete 8-phase multi-intersection scenarios and achieve explainable traffic signal control.
Despite these studies, the application and development of GP to devise effective and explainable traffic signal control strategies are currently still in the early stages of exploration and require further advancements.

\section{Proposed Method}
\label{sec:method}
This section presents the proposed algorithm, namely GPLight+, for the traffic signal control problem.
The overall framework of the proposed GPLight+ is described first, followed by the details of its key components.

\subsection{Overall Framework}
In this work, we propose a new framework called GPLight+.
Similar to the previous work GPLight \cite{liao2024learning}, GPLight+ also operates in a decentralized manner, treating each intersection as a single agent.
Different from GPLight, GPLight+ uses a new GP individual representation along with lane feature permutation.
The overall framework is illustrated in Fig. \ref{fig:framework}.
Firstly, a population containing a set of individuals, each representing a tree-based urgency function, is randomly initialized based on the ramp-half-and-half method \cite{luke2001survey}.
Then the algorithm enters the main loop.
Each individual within the population represents an urgency function, which is subsequently utilized in the simulator to regulate the traffic lights at each intersection. 
After completion of the simulation, the fitness of the individual is determined by recording the average travel time of vehicles.
Upon evaluating all candidate individuals, promising candidate individuals selected by the tournament selection.
Subsequently, the tree-based crossover, tree-based mutation and reproduction with elitism are applied to these individuals, thereby forming a new generation.
This new population will be re-evaluated and selected, repeating the evolution process until the stopping criteria are met.
Finally, the best individual of the final population is returned as the final urgency function.

\subsection{ Urgency Function Representation}

We use the concept of urgency to guide the switching of traffic lights. 
During each traffic light transition, the same phase urgency function will be utilized in calculating the phase urgency value for every phase at every intersection.
At each intersection, the traffic light phase with the highest urgency value should be actuated.
Through this approach, we aim to always choose the signal phase with the highest urgency at each signal transition, thereby reducing waiting times for vehicles at intersections and effectively coordinating traffic flows from different directions.
In this case, the most crucial aspect is to find the most accurate phase urgency function.
To depict the urgency function precisely, multiple new concepts as well as some basic concepts with extension are presented as follows.

\subsubsection{Traffic phase}
In this work, a traffic signal phase refers to a signal light pattern at an intersection to allow several compatible turn movements.
In a standard 8-phase intersection, a single traffic phase $s_i$ typically includes two turn movements, and the set of all phases is represented as:
\newcommand{\rtt}[1]{\rotatebox[origin=c]{#1}{$\Lsh$}}
\begin{gather}
    \mathcal{S} = \left\{ s_0, s_1, s_2, \cdots, s_7 \right \} \\
    = \left\{ \begin{alignedat}{2}
    &\left( \mathcal{T}_0^{\rightarrow\vphantom{\rtt{90}}}, \mathcal{T}_1^{\leftarrow} \right) &,\quad&
    \left( \mathcal{T}_0^{\leftarrow}, \mathcal{T}_1^{\rtt{90}} \right), \\
    &\left( \mathcal{T}_0^{\rtt{270}}, \mathcal{T}_1^{\rightarrow} \right) ,&&
    \left( \mathcal{T}_0^{\rtt{270}}, \mathcal{T}_1^{\rtt{90}} \right), \\
    &\left( \mathcal{T}_0^{\uparrow\vphantom{\rtt{90}}}, \mathcal{T}_1^{\downarrow} \right) ,&&
    \left( \mathcal{T}_0^{\Lsh}, \mathcal{T}_1^{\uparrow} \right), \\
    &\left( \mathcal{T}_0^{\downarrow}, \mathcal{T}_1^{\rtt{180}} \right) ,&&
    \left( \mathcal{T}_0^{\Lsh}, \mathcal{T}_1^{\rtt{180}} \right)
    \end{alignedat} \right\},
\end{gather}
where the superscript to the right of each $\mc{T}$ represents the traffic direction of that TM.

In work \cite{liao2024learning}, a tree-based function, called phase urgency function, is used to compute the priority of each phase.
It takes the features of all lanes relevant to a phase as input and outputs the priority of that phase.
However, this approach has a potential drawback that there is inconsistency in the calculation of phase priority.
Since the features of different phases come from different lanes, it is necessary to impose a certain form of fixed ordering on the input features.
If the input features are not properly handled, it may lead to undesirable situations, such as a variable $x$ of the tree-based function receiving the value from a feature on an incoming lane when calculating the priority of $s_i$, and possibly from a feature on an outgoing lane when calculating the priority of $s_j$.
The work \cite{liao2024learning} addressed this by using a specific feature ordering, but there still remains an unresolved issue of inconsistency.

\begin{figure}[!t]
\centering
\includegraphics[width=0.95\columnwidth]{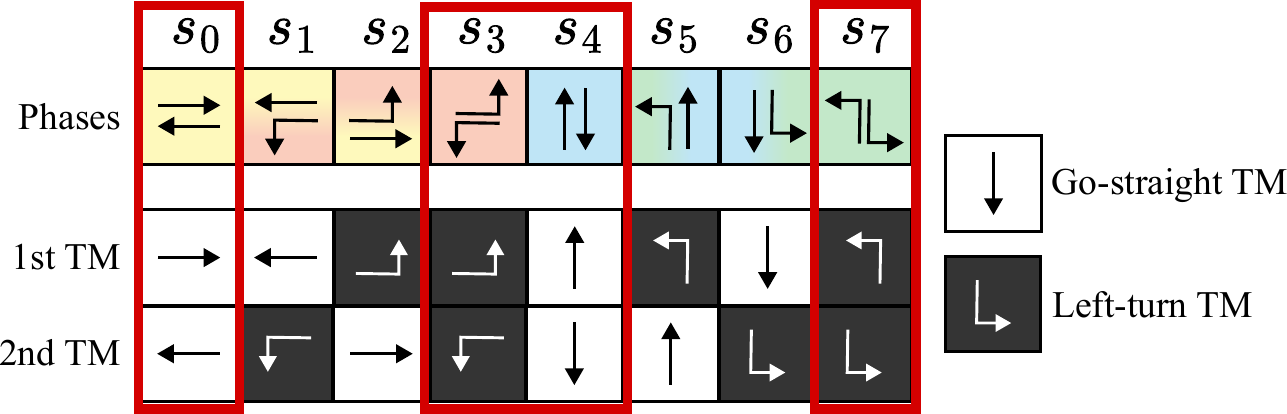}
\caption{Placements of TMs in each traffic signal phase. Each phase consists of two different TMs, and all TMs can be classified into two types: go-straight TMs and left-turn TMs, represented by white and black, respectively.
Since phases $s_0$, $s_3$, $s_4$ and $s_7$ consists of two TMs of the same type, a symmetric phase urgency function is necessary to guarantee the same treatment for two TMs of the same type.}
\label{fig:phase_placements}
\end{figure}

Looking at Fig. \ref{fig:phase_placements}, we can see that there are 8 phases, each containing two turn movements.
It can be also observed that different phases partially share the same TMs.
The features of an entire phase are composed of combinations of features from the TMs it contains.
There are two types of TMs, including left-turn TMs and go-straight TMs.
For phase $s_0$, $s_3$, $s_4$ and $s_7$, which consist of two TMs of the same type, asymmetric phase urgency functions can lead to possible inconsistencies in the urgency calculation of the same type of TMs.
In this case, we propose adding a constraint to the urgency function in this work to ensure consistency and rationality in the calculation.
If a phase urgency function is represented by $\Gamma_{\text{phase}}(\cdot)$, and we assume it has $2\times n$ lane features as input arguments from two TMs, with the first $n$ arguments originating from the first TM and the remaining $n$ arguments coming from the second TM, then we want it to satisfy a special symmetry described in \eqref{eq:constraint}.
\begin{gather}
  \Gamma_{\text{phase}}\left(X_0, X_1\right)  = \Gamma_{\text{phase}}\left(X_1, X_0\right),\label{eq:constraint} \\
  \notag \forall X_0, X_1 \in \mathbb{R}^{n}.
\end{gather}
where $X_0$ and $X_1$ are both ordered sets of size $n$. 
This symmetry is necessary because, when calculating the priority of a phase, the placement of arguments for TMs should not lead to different results.
In the next subsection, we propose a new urgency function representation to impose this constraint on the function during the evolutionary process.

\subsubsection{Symmetric urgency function based on shared subtrees}

\begin{figure}[!t]
\centering
\includegraphics[width=0.8\columnwidth]{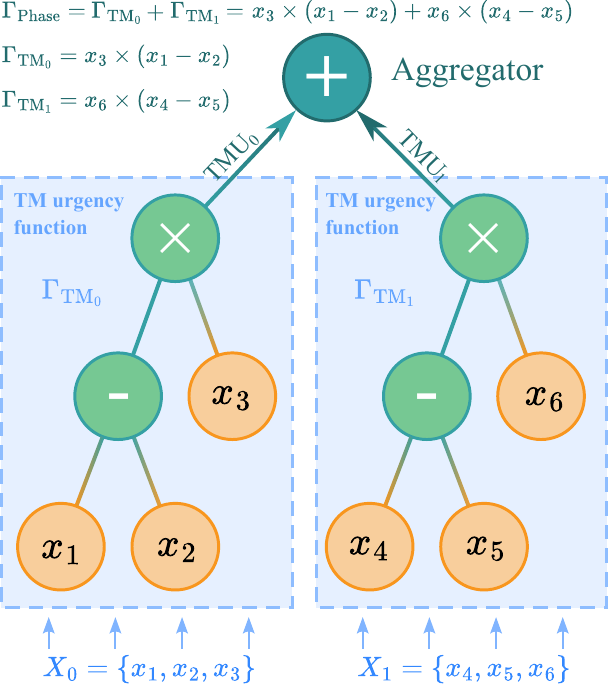}
\caption{A phase urgency function $\Gamma_{\text{Phase}}$ is a tree-shaped function that is composed of two identical TM urgency functions (i.e., $\Gamma_{\text{TM}_0}$ and $\Gamma_{\text{TM}_1}$), where the left and right subtrees of the aggregator are completely isomorphic. Green nodes represent operators, while orange nodes represent variables.}
\label{fig:symmetric_tree}
\end{figure}

In this work, we first evolve a subtree, and then further combine this subtree into a new tree to ensure that the final phase urgency function satisfies the constraints in  \eqref{eq:constraint}.
The schematic of this combination is presented in Fig. \ref{fig:symmetric_tree}.
In this figure, the subtree on the right (shown in the blue area) is a complete copy of the subtree on the left.
The left subtree takes the features (in a predefined order) of the first TM of each phase as input, while the right subtree takes the features (in the same predefined order) of the second TM of each phase as input, without interfering with each other.
Their final outputs will serve as inputs to the aggregation tree.
The aggregator needs to be a symmetric function to satisfy \eqref{eq:constraint} and its output serves as the final output of the entire tree.
This output is the urgency value for the target phase.

The inputs to both the left and right subtrees are features of individual TMs.
For convenience, we assign them specific names related to TMs. 
In this paper, either the left subtree or the right subtree in the figure, is referred to as the TM urgency function, and its output value is called TM urgency (TMU).
We define the phase urgency as the sum of all associated TMUs.
Thus, the aggregator in this work is a simple addition.

Under this design, regardless of the structure of the TM urgency function, the final phase urgency function always satisfies constraint \eqref{eq:constraint}.
Therefore, we can use GP to evolve the optimal TM urgency function without considering that constraint.

\subsubsection{TM involved lanes}

A TM at an intersection, denoted as $\mc{T}_j$ signifies the traffic moving across the intersection from one incoming road $\mc{R}^{\text{in}}$ to an outgoing road $\mc{R}^{\text{out}}$, thus it can be represented as a binary tuple:
\begin{equation}
\mc{T}_j = \left(\mc{R}_j^{\text{in}}, \mc{R}_j^{\text{out}}\right).
\end{equation}

Each road $\mc{R}_r$ comprises several lanes which can be categorized into three types, including left-turn lanes $\mc{L}_r^{\Lsh}$ to allow vehicles on them to turn left, through lanes $\mc{L}_r^{\uparrow}$ to allow vehicles on them to go straight, and turn-right lanes $\mc{L}_r^{\Rsh}$ to allow vehicles on them to turn right.
In this work, we use a triplet to represent a road:

\begin{equation}
    \mc{R}_r= \left(\mc{L}_r^{\Lsh}, \mc{L}_r^{\uparrow}, \mc{L}_r^{\Rsh}\right).
\end{equation}

A traffic turn movement $\mc{T}_j$ involves vehicles on two roads (i.e., $\mc{R}_j^{\text{in}}, \mc{R}_j^{\text{out}}$), but not all lanes on these two roads are relevant to that turn movement. 
A left-turn TM $\mc{T}_j^{\Lsh}$ involves all the lanes on the outgoing lanes and only the left-turn lanes (i.e., $\mc{L}_j^{\Lsh}$) on the incoming road.
Similarly, a go-straight TM $\mc{T}_j^{\uparrow}$ on the incoming road only involves through lanes designated for going straight, that is $\mc{L}_j^{\uparrow}$.
The lanes on the incoming road of a right-turn TM $\mc{T}_i^{\Rsh}$ related to this TM are the right-turn lanes (i.e., $\mc{L}_j^{\Rsh}$).

To represent the above situations conveniently, we introduce a function to denote this relationship of a TM $\mc{T}_j$ and its corresponding incoming lanes $\mc{L}_j^{\text{in}}$:
\begin{equation}
    \mc{L}^{\text{in}}_j = h^{\text{in}}\left( \mc{T}_j \right) 
    = h^{\text{in}}\left( \mc{R}^{\text{in}}, \mc{R}^{\text{out}} \right).
\end{equation}
Therefore, all the lanes involved in a TM can be represented as:
\begin{equation}
    \mc{L}^{\text{TM}}_j = \bigcup \left( \left\{ h^{\text{in}}\left( \mc{T}_j \right) \right\} \cup \mc{R}^{\text{out}}_j \right) .
\end{equation}

\subsubsection{Lane Features}
\begin{figure}[!t]
\centering
\includegraphics[width=0.95\columnwidth]{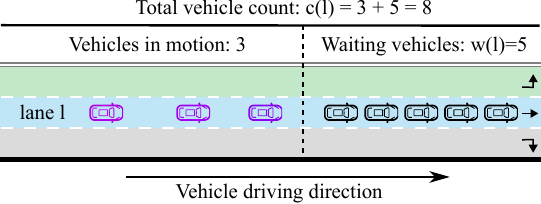}
\caption{Two lane features waiting vehicle number $w(l)$ and vehicle count $c(l)$ in this work.}
\label{fig:lane_features}
\end{figure}

As advancements in vehicular ad-hoc network (VANET) technology \cite{ji2020survey} continue and road infrastructure undergoes continuous improvement, extensive real-time data on important vehicle and road conditions can be gathered and analyzed.
This work utilizes two lane-related features, which is illustrated in Fig. \ref{fig:lane_features}.
These two features have also been extensively employed in previous research \cite{zheng2019diagnosing,zhang2022expression}:
the number of waiting vehicles $w(l)$ and the total number of vehicles $c(l)$ of lane $l$, where $c(l)$ encompasses both vehicles in motion and waiting vehicles.
Therefore, given any turn movement $\mathcal{T}_j$, its set of available features is denoted as:
\begin{equation}
    X_j = \left\{ w\left(l\right), c\left(l\right) \mid l \in \mathcal{L}^{\text{TM}}_j \right\} .
\end{equation}

Since different TMs involve different lanes, it is necessary to ensure consistency when calculating different TMUs.
A predefined feature permutation template is recommended to ensure the consistency.
In this work, we use the template in Fig. S7 to arrange the features $X_j$ of each TM $\mc{T}_j$. 

For each lane-related feature, we input it into the TM urgency function in a specific order to ensure consistency in calculating different TM urgency values, regardless whether it is for the same phase's TMs or different phases' TMs.
We assume that the number of lanes involved in a single TM is $L = |\mc{L}^{\text{TM}}_j|$.
Therefore, the first $L$ inputs of the TM urgency function come from the first lane-related features of different lanes:
\begin{equation}
    \begin{split}
    X_{j;0:L} &= \left[ x_0, x_1, x_2, \cdots, x_{L-1} \right] \\
   &= \left[ \text{feature}_1\left( l \right) \mid l \in \mc{L}^{\text{in}}_j \right] 
   \cup \left[ \text{feature}_1\left( l \right) \mid l \in \mc{L}_j^{\Lsh} \right] \\
   &\cup \left[ \text{feature}_1\left( l \right) \mid l \in \mc{L}_j^{\uparrow} \right]
   \cup \left[ \text{feature}_1\left( l \right) \mid l \in \mc{L}_j^{\Rsh} \right]
\end{split}
\end{equation}
where $X_{j;0:L}$ denotes the ordered set formed by the $a$-th to $b$-th elements of $X_j$. Symbol $[\cdot]$ denotes an ordered set.
The ordering of single-class features follows the sequence: incoming lanes $\mc{L}^{\text{in}}_j$, left-turn lanes $\mc{L}_j^{\Lsh}$, through lanes $\mc{L}_j^{\uparrow}$, and right-turn lanes $\mc{L}_j^{\Rsh}$.
For each class of lanes, they follow the order from left to right in the Euclidean space.
We adopt this ordering sequence of involved lanes for each class of features.
Thus, for each lane-related feature $n$, we can determine its position and ordering in the input of the TM urgency function:
\begin{equation}
\begin{split}
    X_{j;(n-1)L:nL-1} &= \left[ x_{(n-1)L}, x_{(n-1)L+1}, \cdots, x_{nL-1} \right] \\
   &= \left[ \text{feature}_n\left( l \right) \mid l \in \mc{L}^{\text{in}}_j \right] \\
   &\cup \left[ \text{feature}_n\left( l \right) \mid l \in \mc{L}_j^{\Lsh} \right] \\
   &\cup \left[ \text{feature}_n\left( l \right) \mid l \in \mc{L}_j^{\uparrow} \right] \\
   &\cup \left[ \text{feature}_n\left( l \right) \mid l \in \mc{L}_j^{\Rsh} \right] .
\end{split}
\end{equation}

To simplify representation, we use a function $\mc{G}$ to represent the relationship between a TM $\mc{T}_j$ and the ordered set of features $X_j$:
\begin{equation}
    X_j = \mc{G}\left(\mc{T}_j\right) .
    \label{eq_preprocess}
\end{equation}

\subsection{Signal Control via Phase Urgency}
Given that each phase encompasses distinct incoming and outgoing lanes, alongside varying traffic conditions in each lane, where congestion may be prevalent in some while others remain sparsely populated with vehicles. 
Therefore, evaluating the traffic flow conditions in lanes linked to each phase, determining the urgency level for each phase, and opting for the phase with the utmost priority is an intuitive way.
When a phase transition is necessitated at any intersection, the urgency function can be employed to compute the subsequent phase deemed most suitable for activation, as defined by \eqref{eq_selection}.

\begin{equation}
    s^* = \mathop{\arg\max}\limits_{\mc{T}_j \in s_i, s_i \in S}\ \left(
        \sum_{j=0}^1 \text{TMU}_j
    \right),
    \label{eq_selection}
\end{equation}
where
\begin{equation}
    \text{TMU}_j = \Gamma_{\text{TM}}\left( \mc{G}\left(\mc{T}_j\right) \right),
    \label{eq_cal_tmu}
\end{equation}
where $\Gamma_{\text{TM}}(\cdot)$ represents the evolved TM urgency function, which is a tree-based function partially comprising features designed for each TM.
Given a TM urgency function, when a phase transition is needed at an intersection, the urgency value for each turn movement, i.e., TMU can be calculated. 
For the 8 phases, we can identify their corresponding two TMUs and calculate the phase urgency for each phase by summing (since the aggregator is a simple addition) them.
Thus, we can obtain the 8 urgency values for 8 phases, from which the phase with the highest urgency value at that moment can be actuated.

\subsection{GP Components}
In this paper, we utilize genetic programming to evolve the TM urgency functions, referred to as GPLight+.
In GPLight+, each individual in the context of this work is a tree-based function, representing a potential candidate TM urgency function denoted as $\Gamma_{\text{TM}}(\cdot)$.

\begin{table}[h!]
\centering
\caption{Traffic features in this work.}
\renewcommand{\arraystretch}{1.2} 
\setlength{\tabcolsep}{4pt}
\begin{tabular}{|c|c c c c c c c c|} 
\hline
Symbol
& $x_0$ & $x_1$ & $x_2$ & $x_3$ & $x_{4+0}$ 
& $x_{4+1}$ & $x_{4+2}$ & $x_{4+3}$ \\
\hline
Calculation
& $w\left(l\right)$ & $w\left(l\right)$ & $w\left(l\right)$ & $w\left(l\right)$
& $c\left(l\right)$ & $c\left(l\right)$ & $c\left(l\right)$ & $c\left(l\right)$ \\
\hline
Lanes
& $\mc{L}_j^{\text{in}}$ & $\mc{L}_j^{\Lsh}$ & $\mc{L}_j^{\uparrow}$ & $\mc{L}_j^{\Rsh}$ 
& $\mc{L}_j^{\text{in}}$ & $\mc{L}_j^{\Lsh}$ & $\mc{L}_j^{\uparrow}$ & $\mc{L}_j^{\Rsh}$ 
\\ 
\hline
Simplified
& W0 & W1 & W2 & W3 
& C0 & C1 & C2 & C3 \\
\hline
\end{tabular}
\label{table:terminal_set}
\end{table}

The terminal set is crafted to encompass the previously mentioned features associated with a single TM.
In this work, two features waiting vehicle number $w(\cdot)$ and vehicle count $c(\cdot)$ are utilized by GP, thus $n=2$.
In the experimental scenarios, each road consists of three lanes, with one lane dedicated to each type of TM, thus 
$|\mc{L}_j^{\text{in}}| = |\mc{L}_j^{\Lsh}| = |\mc{L}_j^{\uparrow}| = |\mc{L}_j^{\Rsh}| = 1$ and $L = 4$.
For ease of understanding, the explanation of each terminal is presented in Table \ref{table:terminal_set}.
In this representation, it can be clearer that the features $x_{(nL+m_1)}$ and $x_{(nL+m_2)}$ both originate from the same lane.
In this work, we only utilize two types of feature, thus we further adopt simplified symbols to represent these terminals.
In the simplified symbols, each symbol beginning with ``W'' represents the number of waiting vehicles in a specific lane, while each symbol starting with ``C'' denotes the number of vehicles in a lane.
Thus, the terminal set can be further rewritten as:
\begin{equation}
\left\{ \text{W0},\text{W1},\text{W2},\text{W3},\text{C0},\text{C1},\text{C2},\text{C3}, \text{EPM} \right\},
\end{equation}
where EPM is the ephemeral terminal \cite{koza1994genetic} within $[-1, 1]$.
The function set in this work is set as \eqref{eq_func_set}.
\begin{equation}
\left\{+,-,\times,\div\right\}.
\label{eq_func_set}
\end{equation}
Every function operates on two arguments, and the ``$\div$'' function guarantees protected division, returning one in the event of division by zero.

\begin{algorithm}[!t]
\DontPrintSemicolon
\caption{Fitness evaluation/Simulation}
\label{alg_simulation}
\KwIn{TM Urgency function $\Gamma_{\text{TM}}(\cdot)$ and traffic dataset.}
\KwOut{Average travel time of vehicles as the fitness of $\Gamma_{\text{TM}}(\cdot)$.}
$t \leftarrow 0$ \;
\For {each time step $t$}{
    \For{each intersection of the road network}{
        Collect the real-time data at time step $t$ on lanes of the intersection\;
        \For{each phase $s_i$ of the intersection}{
            \For{each TM $\mc{T}_j \in s_i$ of the phase}{
                Obtain $X_j$ for $\mc{T}_j$ by Eq. \eqref{eq_preprocess}\;
                Calculate the $\text{TMU}_j$ by Eq. \eqref{eq_cal_tmu} \;
            }
            Calculate the phase urgency of $s_i$ by summing up all the $\text{TMU}_j$ \;
        }
        
        Identify the current most urgent phase $s^*$ based on Eq. \eqref{eq_selection} \;
        Set the signal phase of the intersection to $s^*$ \;
    }
    Switch the corresponding red and yellow lights based on changes in signal phase at each intersection, and simulate vehicles running until the next signal phase transition. \;
    $t \leftarrow t + 1$ \;
}
Get average travel time from the simulator and return it\; 
\end{algorithm}

Algorithm \ref{alg_simulation} shows the simulation-based evaluation process for a TM urgency function.
The simulation starts by initializing the time step.
The algorithm operates on each intersection of the road network individually. 
For every intersection, it first collects real-time data regarding traffic conditions on the lanes at the current time step $t$.
Then, for each phase $s_i$ of the intersection, it iterates through every TM $\mc{T}_j$ associated with phase $s_i$. 
For each TM, it calculates the features $X_j$ by  \eqref{eq_preprocess} and subsequently computes urgency $\text{TMU}_j$ of $\mc{T}_j$ by \eqref{eq_cal_tmu}.
Once the TMUs for all TMs in a phase are determined, the algorithm computes the phase urgency by aggregating these TMUs.
After evaluating the urgency of each phase, the algorithm identifies the most urgent phase $s^*$ based on \eqref{eq_selection}. 
It then sets the signal phase of the intersection to $s^*$, initiating the corresponding traffic signal changes.
The simulation progresses by switching red and yellow lights based on signal phase changes. 
Vehicle movement continues to be simulated until the next signal phase transition, which takes place during the subsequent time step.
The iterative process persists for multiple time steps until the predetermined simulation duration is reached by the current time reaches.
Once the simulation concludes, the average travel time of vehicles can be computed, offering a crucial metric for assessing the overall effectiveness of a TM urgency function.

\section{Experiments}
\label{sec:experiments}
In this section, we conduct a series of experiments on six traffic flow datasets based on three real-world road networks to evaluate our proposed method. Subsequently, we analyze the experimental results in detail.

\subsection{Datasets}

In this work, we utilize three real-world traffic datasets, which have been frequently employed in previous studies \cite{wu2021efficient,zhang2022expression}: Dongfeng Sub-district in Jinan ($3\times 4=12$ intersections with three traffic flow datasets, namely, $\text{Jinan}_1$, $\text{Jinan}_2$ and $\text{Jinan}_3$), Gudang Sub-district in Hangzhou ($4\times 4=16$ intersections with two traffic flow datasets, namely, $\text{Hangzhou}_1$ and $\text{Hangzhou}_2$), and Upper East Side in Manhattan ($28\times 7=196$ intersections with one traffic flow dataset, namely, $\text{Manhattan}$.). 
All traffic flow datasets have a duration of one hour.
Table S1 presents statistics of all traffic flow datasets and Fig. S2 shows a network illustration of Manhattan.
In all road network datasets, all roads are 3-lane roads, with each lane dedicated to a single TM (turn left, turn right or go straight).
Thus, at each intersection, there are 8 roads and $8 \times 3 = 24$ lanes.

\subsection{Baselines}

\subsubsection{Traditional methods}
\begin{itemize}
    \item \textbf{Random}
    Each phase is assigned a fixed time interval and each time the traffic lights need to switch, a random phase will be actuated regardless of the real-time traffic conditions at the intersection.
    \item \textbf{Fixed-time} 
    Each phase is assigned a fixed time interval and follows a predetermined cyclic schedule, disregarding the real-time traffic conditions at the intersection.
    \item \textbf{Max Pressure (MP)} 
    The state-of-the-art heuristic method \cite{varaiya2013max} in the transportation field.
    This method greedily relieves vehicles on the lanes involved by the signal phase with the highest pressure at each time step and allocates green time accordingly.
\end{itemize}

\subsubsection{Learning-based methods}
\begin{itemize}
    \item \textbf{IDQN} A method \cite{zheng2019diagnosing} that addresses the TSC problem using deep Q-learning, leveraging a concise reward (the average number of waiting vehicles at an intersection) to guide the learning process.
    \item \textbf{FRAP} A method \cite{zheng2019learning} designs a novel neural network model that incorporates phase relationships, aiming to effectively manage possible unbalanced traffic flows.
    \item \textbf{PressLight}
    A DRL-based TSC method \cite{wei2019presslight} that train different traffic signal control policies for different intersections. 
    This method designs the reward based on pressure theory instead of relying on a heuristically designed reward.

    \item \textbf{MPLight} 
    A DRL-based TSC method \cite{chen2020toward} that shares a similar control approach to our work. It utilizes a decentralized control strategy where agents, each responsible for managing an intersection, share neural network parameters. 
        It integrates FRAP as the policy model and employs a theoretically supported reward design \cite{wei2019presslight}. 
        This decentralized, parameter-sharing control approach closely aligns with the framework of our proposed method.

    \item \textbf{GPLight} 
    A method that uses GP to learn traffic signal control policies \cite{liao2024learning}.
    In this method, the phase urgency function lacks the symmetric constraint and is directly evolved by GP.
\end{itemize}

Notably, all traditional methods and IDQN use our own versions and all the other RL-based methods use the implementations provided by the authors of the original papers.

\subsection{Settings}

\begin{table}[!t]
\setlength{\tabcolsep}{22pt} 
\renewcommand{\arraystretch}{1.0} 
\caption {Parameter settings.}
\centering
\begin{tabular}{cc}
\hline
\hline
Parameter      & Value  \\ 
\hline
Population size   & 100 \\ 
The number of generations & 51 \\ 
Method of initialization & ramped-half-and-half\\ 
Initial minimum depth & 3  \\ 
Maximum depth & 6 \\
Elitism & 1 \\
Parent selection  & tournament selection \\
Tournament size & 3 \\
Crossover rate & 90\% \\
Mutation rate  &  10\% \\ 
\hline
\hline
\end{tabular}
\begin{tablenotes}
  \item \textit{Initial minimum depth} refers to the minimum tree depth during initialization.
  \item \textit{Maximum depth} is the maximum depth allowed for any tree in the population at any generation.
  \item \textit{Elitism} refers to the number of top-performing individuals preserved unchanged for the next generation.
\end{tablenotes}
\label{table_parameters}
\end{table}

We conduct our experiments using Cityflow\footnote{https://cityflow-project.github.io}, an open-source traffic simulator specifically developed for large-scale traffic signal control scenarios.
After the traffic data is input into the simulator, vehicles move toward their destinations based on the environmental settings.
The simulator furnishes local information within an intersection to the traffic signal control method, which subsequently executes the corresponding traffic signal phases.
As per convention \cite{wei2019presslight}, each green signal is followed by a three-second yellow signal as well as a two-second all-red interval to clear the intersection.
Each fitness evaluation of GP is a 60-minute simulation.
We set the minimum action duration to 10-second \cite{wei2019colight} which is consistent when establishing the baseline.
Based on recommendations from previous work \cite{liao2024learning}, the parameters of GPLight+ are presented in Table \ref{table_parameters}.

\subsection{Metrics}

In this work, we utilize metrics such as \textit{average travel time} (ATT), \textit{average queue length} (AQL), and \textit{network throughput} (NT) of vehicles to assess the effectiveness of various policies implemented by traffic signal control.

\subsubsection{Average travel time (ATT)}

The travel time of each vehicle (the smaller the better) is defined as the time difference, measured in seconds, between when the vehicle enters and exits a transportation area. 
The average travel time of all vehicles in a traffic network is the most commonly used metric \cite{liao2022combining} to evaluate the performance of TSC and serves as the fitness in the evaluations for GP-based methods.

\subsubsection{Average queue length (AQL)} 

The average queue length (the smaller the better) refers to the average number of vehicles queued up and waiting within the road network.

\subsubsection{Network throughput (NT)} 
The network throughput of vehicles (the bigger the better) refers to the total number of vehicles that have entered and exited the road network \cite{liao2023towards,liao10816321}.

\subsection{Performance Comparison}

\makeatletter
\newcommand*{\txtoverline}[1]{$\overline{\hbox{#1}}\m@th$}
\makeatother

\newcommand*{\comparetablenotes}{%
\item Columns represent different datasets, while the rows represent different algorithms.
\item ATT, AQL, and NT are the three distinct evaluation metrics, that is average travel time (the smaller the better), average queue length (the smaller the better) and network throughput (the larger the better) of vehicles.
\item The symbols \textit{avg.} and \textit{std.} represent the average values and standard deviation, respectively, of the results from 30 independent runs. Symbol \textit{best} represents the best values (i.e., the minimum values of ATT/AQL and the maximum values of NT) of the results from 30 independent runs.
\item The symbols "+/-/$\approx$" indicate whether the corresponding metric is significantly worse than, better than or statistically comparable to the proposed GPLight algorithm, based on the Wilcoxon rank-sum test at a significance level of 0.05 with Bonferroni correction. 
\item The best values (\textit{best} and \textit{avg.}) among all the compared methods of ATT, AQL and NT in each scenario are highlighted;
}

\begin{table*}[!t]
\renewcommand{\arraystretch}{1.1} 
\setlength{\tabcolsep}{5.5pt}
\caption {The results of 30 independent runs of different metrics in $\text{Jinan}_1$, $\text{Jinan}_2$ and $\text{Jinan}_3$.}
\centering
\begin{tabular}{c|l|c|c|c|c|c|c|c|c|c}
\hline\hline
\multicolumn{2}{c|}{\multirow{2}{*}{Method}} 
&\multicolumn{3}{c|}{$\text{Jinan}_1$} &\multicolumn{3}{c|}{$\text{Jinan}_2$} &\multicolumn{3}{c}{$\text{Jinan}_3$} \\ \cline{3-11}
\multicolumn{2}{c|}{} &\textbf{ATT} &AQL &NT &\textbf{ATT} &AQL &NT &\textbf{ATT} &AQL &NT \\ \hline


Random      &avg. &735.3125 &57.0861 &5238.0 &634.6188 &35.0127 &4088.0 &671.2037 &44.6171 &4668.0 \\ \hline
Fixed-time \cite{webster1958traffic}  &avg. &542.4370 &36.6123 &5676.0 &431.4845 &17.9345 &4245.0 &481.5946 &25.7690 &5063.0 \\ \hline
Max-pressure \cite{varaiya2013max}&avg. &374.5641 &19.5648 &6101.0 &328.9629 &8.1778 &4306.0 &332.9403 &12.1396 &5356.0 \\ \hline\hline


&best 
&327.6437 &12.2880 &6162.0 &304.5425 &5.5382 &4319.0 &303.7150 &8.1220 &5379.0 
\\MPLight \cite{chen2020toward} &avg. 
&341.1644(+) &14.5163(+) &6146.1(-) &309.3685(+) &5.9697(+) &4315.5(+) &314.9360(+) &9.5896(+) &5368.4(+) 
\\&std. 
&7.9328 &1.2536 &6.9 &2.3608 &0.2352 &2.0 &7.9433 &1.0545 &5.9 
\\ \hline

&best 
&338.8920 &14.2826 &6155.0 &300.5310 &5.1296 &4320.0 &312.0277 &9.2053 &5381.0    
\\FRAP \cite{zheng2019learning} &avg. 
&343.0691(+) &14.8591(+) &6147.3(-) &303.2455(+) &5.3727(+) &4315.9(+) &316.6519(+) &9.7945(+) &5369.5(+) 
\\&std. 
&2.3429 &0.3489 &5.1 &1.4596 &0.1427 &1.8 &2.7451 &0.3418 &4.3 
\\ \hline

&best 
&324.1749 &12.0581 &\textbf{6176.0} &299.1214 &5.0706 &4319.0 &295.0559 &7.2329 &5385.0    
\\IDQN \cite{zheng2019diagnosing} &avg. 
&334.0328(+) &13.6807(+) &6140.1($\approx$) &301.0628(+) &5.2492(+) &4316.1(+) &297.5135(+) &7.5253(+) &5377.3(+)  
\\&std. 
&8.8895 &1.3760 &36.7 &1.0837 &0.0877 &2.0 &1.0450 &0.1418 &4.4 
\\ \hline

&best 
&\textbf{313.8327} &\textbf{10.7035} &\textbf{6176.0} &297.2282 &5.0947 &4318.0 &292.9701 &7.1532 &5386.0  
\\PressLight \cite{wei2019presslight} &avg. 
&\textbf{315.7515}(-) &\textbf{11.0108}(-) &\textbf{6163.7}(-) &306.0561(+) &6.1158(+) &4303.1(+) &294.9840(+) &7.4287(+) &5373.9(+)  
\\&std. 
&1.2994 &0.1908 &7.5 &38.7354 &4.4130 &56.5 &1.1401 &0.1424 &5.3
\\ \hline

&best 
&328.2550 &12.0394 &6158.0 &279.0039 &2.8991 &4320.0 &288.1649 &5.7900 &5386.0 
\\GPLight \cite{liao2024learning} &avg. 
&337.8758(+) &13.2245(+) &6123.4($\approx$) &280.6863(+) &3.0370(+) &4317.3(+) &293.2043(+) &6.3835(+) &5379.0(+) 
\\&std. 
&5.8481 &0.7029 &31.4 &0.9310 &0.0695 &1.6 &3.4909 &0.3776 &3.7
\\ \hline

&best 
&323.6775 &11.2648 &6156.0 &\textbf{277.6662} &\textbf{2.8046} &\textbf{4322.0} &\textbf{284.2270} &\textbf{5.4484} &\textbf{5388.0} 
\\GPLight+ &avg. 
&329.1386 &12.0057 &6124.7 &\textbf{278.9359} &\textbf{2.9079} &\textbf{4319.1} &\textbf{287.1021} &\textbf{5.7154} &\textbf{5382.8} 
\\&std. 
&2.6028 &0.3152 &20.2 &0.9855 &0.0811 &1.4 &1.5910 &0.1549 &3.1
\\ \hline
\hline
\end{tabular}
\begin{tablenotes}
     \comparetablenotes
\end{tablenotes}
\label{table:performance_first3}
\end{table*}

\begin{table*}[!t]
\renewcommand{\arraystretch}{1.1} 
\setlength{\tabcolsep}{5.5pt}
\caption {The results of 30 independent runs of different metrics in $\text{Hangzhou}_1$, $\text{Hangzhou}_2$ and $\text{Manhattan}$.}
\centering
\begin{tabular}{c|l|c|c|c|c|c|c|c|c|c}
\hline\hline
\multicolumn{2}{c|}{\multirow{2}{*}{Method}} 
&\multicolumn{3}{c|}{$\text{Hangzhou}_1$} &\multicolumn{3}{c|}{$\text{Hangzhou}_2$} &\multicolumn{3}{c}{$\text{Manhattan}$} \\ \cline{3-11}
\multicolumn{2}{c|}{} &\textbf{ATT} &AQL &NT &\textbf{ATT} &AQL &NT &\textbf{ATT} &AQL &NT \\ \hline


Random      &avg. &730.9946 &16.4264 &2400.0 &624.8446 &24.5392 &3875.0 &1595.9515 &21.5014 &7623.0 \\ \hline
Fixed-time \cite{webster1958traffic} &avg. &585.1532 &11.7201 &2600.0 &566.7983 &18.4193 &4214.0 &1557.4579 &21.2365 &7825.0 \\ \hline
Max-pressure \cite{varaiya2013max}&avg. &365.0634 &3.5972 &2928.0 &446.9059 &14.8858 &5552.0 &1335.7877 &17.3380 &9745.0 \\ \hline\hline


&best 
&327.3208 &1.4457 &2933.0 &416.5344 &11.8872 &5711.0 &1201.2206 &12.8392 &10182.0 
\\MPLight \cite{chen2020toward} &avg. 
&329.8745(+) &1.5877(+) &2930.2(+) &423.1737(+) &12.7109(+) &5558.3(-) &1240.7652(+) &14.7876(+) &9596.3(+) 
\\&std. 
&1.5704 &0.0862 &1.1 &4.2424 &0.4023 &86.8 &32.9315 &1.0636 &310.7 
\\ \hline

&best 
&322.5049 &1.2050 &2934.0 &410.2115 &10.7983 &5744.0 &1217.9002 &14.3075 &9918.0 
\\FRAP \cite{zheng2019learning} &avg. 
&324.8163(+) &1.3190(+) &2931.1(+) &417.7774(+) &11.6963(+) &5638.9(+) &1273.1518(+) &15.7991(+) &9339.9(+) 
\\&std. 
&0.9707 &0.0523 &1.5 &3.1446 &0.4515 &48.4 &35.3047 &1.0406 &298.7 
\\ \hline

&best 
&324.2947 &1.2780 &2933.0 &406.4513 &8.9266 &5769.0 &1191.2740 &12.8457 &9924.0   
\\IDQN \cite{zheng2019diagnosing} &avg. 
&324.8687(+) &1.3151(+) &2931.3(+) &412.7171(+) &10.7329(+) &5588.2($\approx$) &1233.0165(+) &14.4369(+) &9392.5(+) 
\\&std. 
&0.3929 &0.0221 &1.2 &4.5918 &0.6180 &84.4 &31.1538 &1.0588 &273.0  
\\ \hline

&best 
&327.3315 &1.5073 &2933.0 &\textbf{392.9712} &8.1948 &\textbf{5782.0} &1152.7073 &12.5499 &10222.0   
\\PressLight \cite{wei2019presslight} &avg. 
&328.8984(+) &1.5890(+) &2929.8(+) &\textbf{396.9998}(-) &8.8858(+) &\textbf{5699.4}(-) &1239.6213(+) &15.2279(+) &9642.2(+)  
\\&std. 
&0.8389 &0.0439 &1.7 &2.9578 &0.5064 &49.5 &38.8961 &1.2356 &197.6
\\ \hline

&best 
&312.6889 &0.7418 &\textbf{2935.0} &401.1413 &7.0682 &5612.0 &1197.1586 &12.6248 &10292.0 
\\GPLight \cite{liao2024learning} &avg. 
&314.9438(+) &0.8222(+) &2932.9(+) &404.6308(-) &8.0658($\approx$) &5462.7(-) &1231.2341(+) &13.9919(+) &9813.8(+) 
\\&std. 
&1.2608 &0.0445 &1.0 &2.2085 &0.4696 &61.6 &20.4218 &0.6605 &175.3
\\ \hline

&best 
&\textbf{311.8877} &\textbf{0.7102} &\textbf{2935.0} &404.1919 &\textbf{6.8243} &5502.0 &\textbf{1131.4515} &\textbf{11.0678} &\textbf{10391.0} 
\\GPLight+ &avg. 
&\textbf{312.9909} &\textbf{0.7497} &\textbf{2933.6} &410.1948 &\textbf{7.9634} &5388.6 &\textbf{1181.4160} &\textbf{12.5822} &\textbf{10003.2} 
\\&std. 
&1.4234 &0.0531 &0.8 &3.6148 &0.4679 &69.3 &22.0707 &0.7128 &291.4
\\ \hline
\hline
\end{tabular}
\begin{tablenotes}
     \item See the footnotes of Table \ref{table:performance_first3}.
\end{tablenotes}
\label{table:performance_last3}
\end{table*}

To verify the effectiveness of the proposed GPLight+, 
we evaluated the average travel time of vehicles for all compared algorithms using the CityFlow engine \cite{zhang2019cityflow}. 
The results of the comparison, conducted over 30 independent runs across 6 datasets, are presented in Table \ref{table:performance_first3} and Table \ref{table:performance_last3}. 
Specifically, the experiment results on $\text{Jinan}_1$, $\text{Jinan}_2$ and $\text{Jinan}_3$ are presented in Table \ref{table:performance_first3} and experiment results on $\text{Hangzhou}_1$, $\text{Hangzhou}_2$ and $\text{Manhattan}$ are presented in Table \ref{table:performance_last3}.
Columns represent different datasets, while the rows represent different algorithms.
The symbols "+/-/$\approx$" indicate whether the corresponding metric is significantly worse than, better than or statistically comparable to the proposed GPLight algorithm, based on the Wilcoxon rank-sum test at a significance level of 0.05 with Bonferroni correction. 
In Table \ref{table:performance_first3} and Table \ref{table:performance_last3}, the best values (the minimum values of ATT/AQL and the maximum values of NT) and mean values among all compared algorithms are marked highlighted.
Specifically, the minimum values and mean values of ATT and AQL, the maximum values and mean values of NT are highlighted.

Regarding learning-based methods, we observe that the proposed GPLight+ exhibits the best performance among all the compared algorithms in most datasets, based on three evaluation metrics.
PressLight, the baseline that trains different TSC policies for different intersections, outperforms all other algorithms in the two small scenarios.
This demonstrates the potential advantage of customizing policies for different intersections in some scenarios.
However, this approach ultimately leads to a collection of specialized policies tailored to each intersection, which significantly limits its transferability across different scenarios, especially those with varying road networks.
Both GPLight and GPLight+ use average travel time of vehicles as the evaluation fitness.
In terms of ATT, GPLight+ performs significantly better than GPLight on 5 out of 6 datasets, especially in large-scale scenarios, i.e. Manhattan, with a considerable reduction in ATT.
This demonstrates the effectiveness of the design in maintaining symmetry for the phase urgency function.

As can be seen from the tables, almost all traditional methods, such as Random, Fixed-time, and Max-pressure, lag behind the learning-based methods.
From the results, it can be observed that the Random control strategy performs poorly.
This indicates that an inadequate traffic signal control strategy can lead to very poor vehicle flow conditions, resulting in congestion and higher average travel times for vehicles.
Based on human design, the Fixed-time method achieves a significant performance improvement compared to the Random method.
However, its performance remains unsatisfactory, indicating that a static traffic signal plan is not well-suited for dynamic traffic conditions in real-world traffic scenarios.
Max-pressure method shows significant improvement compare to other traditional algorithms, but it still falls short of learning-based methods.
This shortcoming is attributed to its heavy reliance on oversimplified assumptions. In complex traffic scenarios, such simplistic assumptions can easily lead to local optima.

Additionally, in most cases, metrics ATT, AQL, and NT exhibit the same trend, algorithms that can achieve the lowest ATT often also yield better AQL and NT.
In GPLight+, only ATT is considered as the fitness, meaning that the objective of the algorithm is solely to minimize the average travel time of vehicles.
However, experimental results show that GPLight+ achieves the highest throughput in most scenarios and obtains the smallest queue length in all scenarios.
This indicates that the proposed algorithm reduces the average travel time without sacrificing other metrics, leading to an overall improvement in network efficiency.

\begin{figure}[!t]
\centering
\includegraphics[width=0.9\columnwidth]{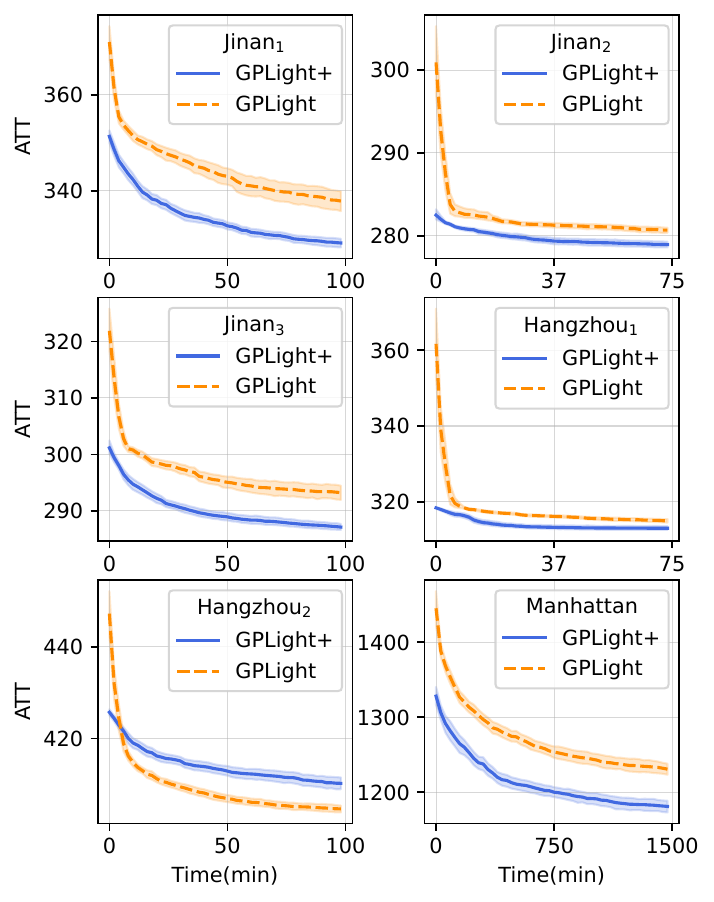}
\caption{Convergence curves of the optimal value during the evolution process of GPLight and GPLight+, with the shadowed regions representing the 95\% confidence interval.}
\label{fig:convergence}
\end{figure}

Fig. \ref{fig:convergence} shows the convergence curves of the performance of the best individual in the population during the evolution process of GPLight and GPLight+, with the shadowed regions representing the 95\% confidence interval.
The vertical axis represents the average travel time of vehicles, while the horizontal axis represents the evolution time in minutes.
From Fig. \ref{fig:convergence}, it can be observed that after introducing the symmetry constraint, GPLight+ can always produce a better urgency function during initialization compared to GPLight in all scenarios.
In five out of the six scenarios, GPLight+ consistently demonstrates better data efficiency than GPLight.
In $\text{Hangzhou}_2$, GPLight+ performs slightly worse than GPLight. 
There are two possible reasons for this: 1) the introduction of the symmetry constraint, to some extent, restricts the search space of phase urgency functions, and 2) the simple and fixed aggregator potentially contributes to the performance limitations of phase urgency functions.
Fortunately, apart from $\text{Hangzhou}_2$, GPLight+ significantly outperforms GPLight in all other scenarios, especially in large-scale settings, with an average travel time reduction of several tens of seconds.
There are two potential benefits from the proposed symmetric phase urgency function.
\begin{enumerate}
    \item Since the symmetric phase urgency function shares the left and right subtrees, it is sufficient to evolve only one subtree. The advantage of this approach is that the size of the terminals set for the evolved subtree is reduced to half of the original, which effectively compresses the search space of GP.
        This benefit can be observed in Fig. \ref{fig:convergence}, where GPLight+ can generate significantly better phase urgency functions than GPLight during the population initialization.

    \item The symmetry constraint can potentially guide GP to evolve urgency functions in a better direction.
    The heuristic space of the phase urgency function in GPLight+ is a subset of that in GPLight. 
    This means that GPLight also has the chance to discover the symmetric phase urgency functions proposed by GPLight+.
    However, as shown in Fig. \ref{fig:convergence}, GPLight+ is able to find better urgency functions in most scenarios that GPLight cannot discover.
    This suggests that GPLight has the chance to find, but struggles to locate, the urgency functions that GPLight+ can search for, which means the symmetry constraint helps filter out most of the low-quality urgency functions.
\end{enumerate}
Furthermore, according to the time in the horizontal axis, it can be seen that even in large-scale scenarios, our method is capable of obtaining effective urgency functions within an acceptable evolution time (about 25 hours on 20 CPUs with the model of Intel(R) Xeon(R) CPU E5-2695 v4 @ 2.10GHz).

\subsection{Explanation of the Learned Urgency Functions}

The advancement of symbolic expressions emerges as a significant benefit of GP, often simplifying the comprehension and explanation of evolved rules for humans. 
To explore the significance of features, we initially illustrate the average occurrence frequency of each terminal in the optimal solutions across six datasets of Fig. S3.
It is evident from the figure that W0 and C0, which respectively represent the number of waiting vehicles and the total vehicle count on the incoming lane, appear with notably high frequencies.
If a phase $s_i$ is actuated, vehicles on the incoming lane $l \in \mc{L}_j^{\text{in}}$ of every $\mc{T}_j$ are the ones most directly affected, while vehicles on other outgoing lanes $l \in \mc{L}_j^{\Lsh} \cup \mc{L}_j^{\uparrow} \cup \mc{L}_j^{\Rsh}$ are only indirectly influenced by the activation of that phase.
In most scenarios, the features on the outgoing lanes tend to have relatively low occurrence frequency. This indicates that the traffic conditions on the incoming lanes are more important for traffic light signal control decisions than the vehicle status on the outgoing lanes.
This also suggests that our new method can automatically detect important features and utilize the more important ones to construct urgency functions.

Additionally, we can also observe that, among the outgoing lanes, the number of vehicles in the right-turn lane $l \in \mc{L}_j^{\Rsh}$, denoted as C3, also has a high occurrence frequency.
This suggests that the vehicle status in the right-turn lane of outgoing lanes contributes significantly to the urgency of the phase.
In our experimental setup, vehicles making right turns are not subject to traffic light signals and C3 is often negatively correlated with the urgency of a phase in the extracted urgency functions.
Based on these observations, we argue a possible reason for this phenomenon is that if the right-turn lane in the outgoing lanes is congested, it indicates relatively severe congestion ahead.
Therefore, vehicles should not be allowed to continue entering the outgoing lanes, resulting in a low urgency value for the corresponding phase.

To examine the explainability of the traffic signal control strategy, we aim to analyze an example of the urgency function evolved by GPLight+.
Fig. S8 presents the tree structure of a TM urgency function trained by GPLight+ on $\text{Hangzhou}_1$ and its corresponding mathematical expression is:
\begin{equation}
\Gamma_{\text{TM}} = \lambda\text{W0} + \left(1-\lambda\right)\text{C0},
\end{equation}
where $\lambda \approx$ 0.9 is a constant.
From this example TM urgency function, we can find that only features on the incoming lane $l \in \mc{L}_j^{\text{in}}$ are considered by $\Gamma_{\text{TM}}$.
In this case, each phase with the potential maximum queue length will be allowed to proceed, which can effectively reduce the queue length on the incoming lanes.
This urgency function also aligns well with the results in Table \ref{table:performance_last3}, where GPLight+ achieves the smallest average queue length (AQL).
In cases of low traffic flow, allowing vehicles on incoming lanes with the longest queue to cross the intersection usually does not lead to downstream congestion, and this indeed helps to effectively reduce the queue length at the intersection.
Therefore, we argue that focusing solely on the upstream traffic conditions is potentially sufficient to improve the transportation efficiency in relatively small-scale scenarios where cross-intersection congestion is not likely to occur.

\subsection{Model Generalization}

\begin{figure*}
    \centering

    \subfloat[Models trained on $\text{Jinan}_1$.]{ \includegraphics[width=0.32\textwidth]{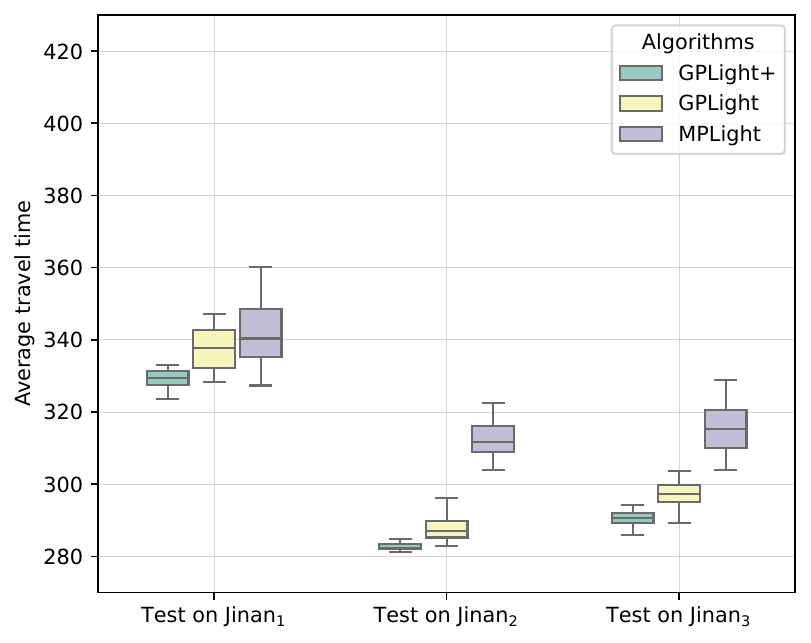} }
    \subfloat[Models trained on $\text{Jinan}_2$.]{ \includegraphics[width=0.32\textwidth]{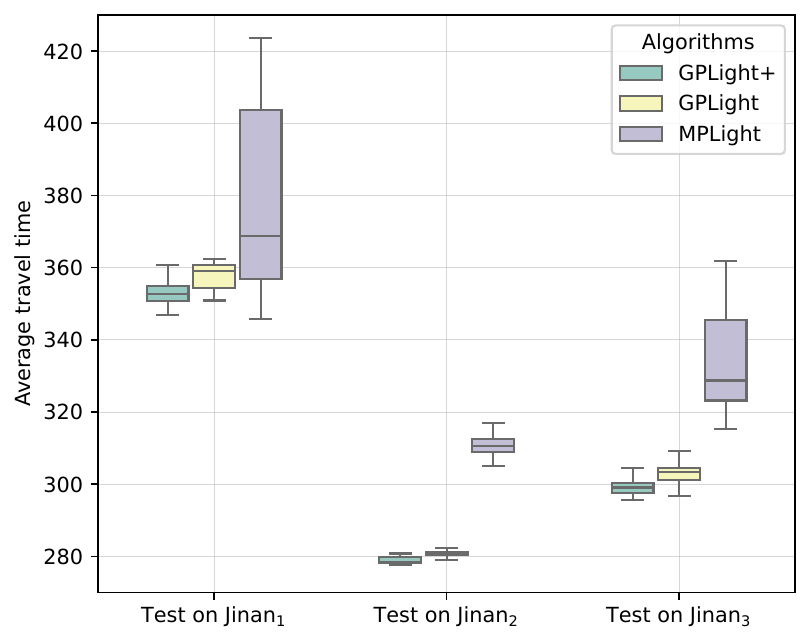} }
    \subfloat[Models trained on $\text{Jinan}_3$.]{ \includegraphics[width=0.32\textwidth]{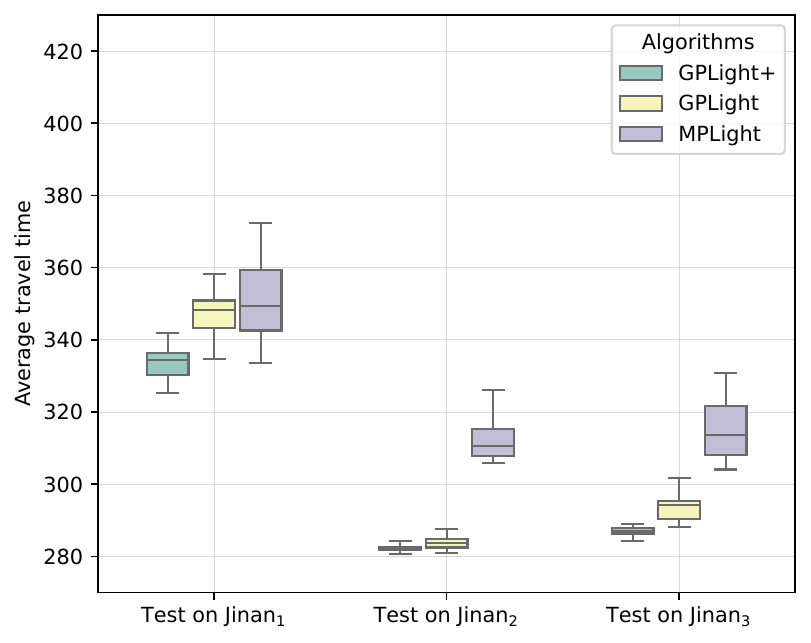} }

    \caption{Average travel time of transfer from different learning-based methods.}
    \label{fig:transferability_analysis}
\end{figure*}

To evaluate the generalization ability of the proposed GPLight+, cross-validation of three learning-based methods on three different datasets from the Jinan road network is preformed.
Fig. \ref{fig:transferability_analysis} reports the average travel time of transfer from different learning-based methods.
Specifically,
Fig. \ref{fig:transferability_analysis} (a) presents the testing results of the model, trained on $\text{Jinan}_1$, when directly applied to $\text{Jinan}_2$ and $\text{Jinan}_3$.
Fig. \ref{fig:transferability_analysis} (b) presents the testing results of the model, trained on $\text{Jinan}_2$, when directly applied to $\text{Jinan}_1$ and $\text{Jinan}_3$.
Fig. \ref{fig:transferability_analysis} (c) presents the testing results of the model, trained on $\text{Jinan}_3$, when directly applied to $\text{Jinan}_1$ and $\text{Jinan}_2$.
From the results provided in the three subplots, we can see that GPLight+ has a smaller variance in average travel time of vehicles compared to other learning-based methods.
GPLight exhibits greater variance compared to GPLight+ in a few scenarios, indicating that satisfying the symmetry of the phase urgency function can effectively enhance its generalizability.
Compared to GP-based methods, MPLight method exhibits relative poor performance and has a larger variance across all scenarios.
Besides, by comparing the same scenario across different subfigures of Fig. \ref{fig:transferability_analysis}, it is evident that GP-based methods exhibit a minimal performance gap between the training and testing datasets.
Based on these observations, we argue that a concise urgency function evolved using GP potentially has better generalizability compared to deep neural network-based models.

\begin{figure}[!t]
\centering
\includegraphics[width=\columnwidth]{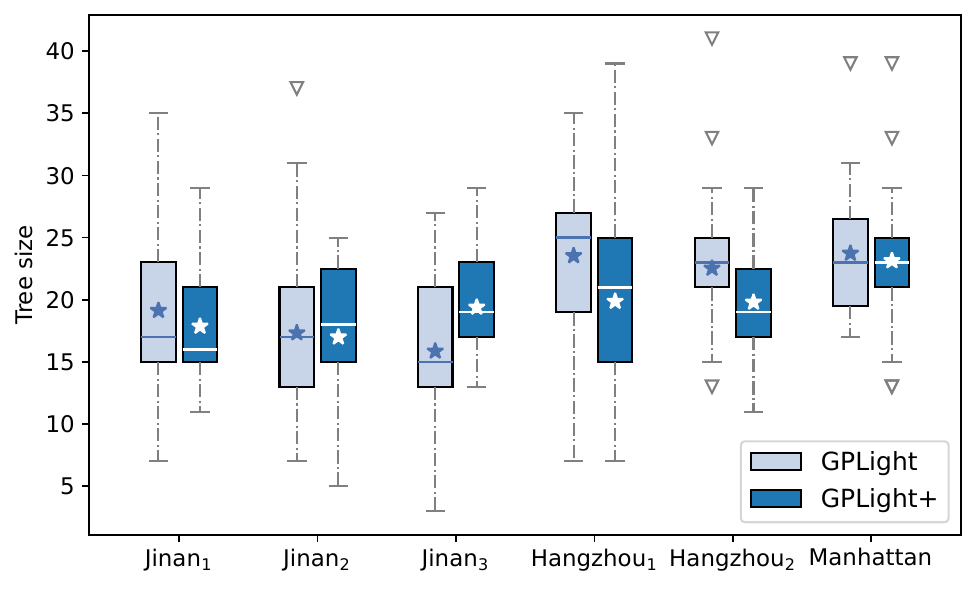}
\caption{Model sizes of the best individual according to 30 independent runs across different scenarios.}
\label{fig:tree_size}
\end{figure}

\subsection{Resource Consumption}

A significant advantage of our method compared to neural network-based reinforcement learning is that our approach can derive an explainable and concise traffic light control strategy.
In order to evaluate resource consumption, We analyze the traffic light control strategy model of each algorithm based on two criteria: storage and computation, which together determine the minimum configuration required for the device.
We utilize two metrics: floating-point operations (FLOPs) to reflect the computation time and model parameter size to reflect memory consumption.
Both FLOPs and parameter size are quantifiable \cite{xing2022tinylight} and independent of irrelevant conditions such as optimization flags on compilers and clock rate.

\begin{figure}[!t]
\centering
\includegraphics[width=\columnwidth]{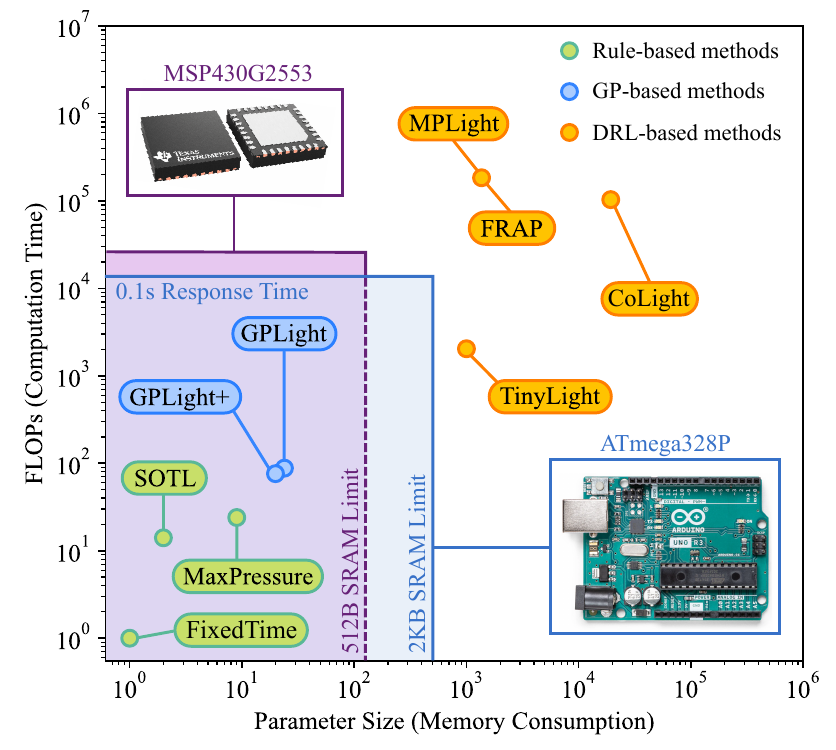}
\caption{Log-log plot for resource consumption of each traffic signal control strategy.}
\label{fig:resource_consumption}
\end{figure}

Fig. \ref{fig:resource_consumption} presents the FLOPs and parameter sizes of traffic signal control strategy from each algorithm from $\text{Hangzhou}_1$ (the same conclusion can be drawn in other scenarios, the model sizes of the best TM urgency functions according to 30 independent runs across six scenarios are presented in Fig. \ref{fig:tree_size}).
To make these values more intuitive, we plot in Fig. \ref{fig:resource_consumption} the borderlines of the RAM and 0.1 second response time from two famous microcontrollers according to TinyLight \cite{xing2022tinylight}.

Compared to traditional methods in the transportation field, GP-based methods have slightly larger scales of parameter size and computation. 
Nevertheless, they remain well within the capabilities of two typical microcontrollers, while delivering superior performance compared to traditional methods.
Comparing with other DRL-based methods, the resource consumption of traffic light control strategies obtained by GP-based methods is significantly lower than that of strategy models based on neural networks.
Our control strategy is based on simple arithmetic operations, it can be directly executed on resource-constrained microcontrollers like the widely-used \cite{jyothi2016smart} ATmega328P\footnote{https://store.arduino.cc/products/arduino-uno-rev3/} and MSP430G2553\footnote{https://www.ti.com/product/MSP430G2553} \cite{gu2024pi}.
This eliminates the need for quantization steps \cite{han2015deep} typically required by neural networks, providing a significant advantage for deployment on edge devices and in practical traffic applications.

\section{Conclusions}
\label{sec:conclusion}
In this paper, we propose a new learning-based method of using GP to evolve traffic signal control strategies, in which the phase urgency function with a symmetric constraint is developed to yield a determination of which signal phase is more urgent.
The experimental results suggest that the proposed symmetry constraint can significantly improve the effectiveness of phase urgency function.
Additionally, GP-based methods exhibit better generalizability than other learning methods, and the final traffic signal control strategy is human-understandable and has lower resource consumption.
This makes our method have higher potential in application of real-world traffic scenarios.
 
This work opens up some promising directions for future research.
The traffic signals at multiple intersections are managed in a distributed manner in this work using the introduced urgency function. 
A promising future direction involves incorporating communication mechanisms to further enhance the algorithm's performance.
Leveraging the unique advantages of GP to evolve multi-objective traffic signal control strategies is also a highly promising research direction.
Besides, this work utilizes a simple addition as the aggregator for phase urgency function, which could lead to potential performance limitations. A promising direction can be exploring or evolving more advanced symmetric aggregators.
Additionally, this work focuses on intersections with dedicated lanes, which is a common design choice for high-traffic intersections. 
Future research could investigate more suitable urgency function representation for specialized intersections with shared lanes.


\bibliographystyle{IEEEtran}
\bibliography{IEEEabrv,gplight_plus_ref}

\end{document}